# Language Without Words: A Pointillist Model for Natural Language Processing

Peiyou Song*, Anhei Shu†, David Phipps‡, Mohit Tiwari§,
Dan S. Wallach†, Jedidiah R. Crandall*, George F. Luger*
*Univ. of New Mexico, Dept. of Computer Science, Email: {peiyou, crandall, luger}@cs.unm.edu
†Rice Univ., Dept. of Computer Science, Email: {as43, dwallach}@rice.edu
‡Bowdoin College, Email: dphipps@bowdoin.edu
§Univ. of California, Berkeley, Dept. of EECS, Email: tiwari@eecs.berkeley.edu

***Abstract*—This paper explores two separate questions: Can we perform natural language processing tasks without a lexicon?; and, Should we? Existing natural language processing techniques are either based on words as units or use units such as grams only for basic classification tasks. How close can a machine come to reasoning about the meanings of words and phrases in a corpus without using any lexicon, based only on grams?**

**Our own motivation for posing this question is based on our efforts to find popular trends in words and phrases from online Chinese social media. This form of written Chinese uses so many neologisms, creative character placements, and combinations of writing systems that it has been dubbed the "Martian Language." Readers must often use visual queues, audible queues from reading out loud, and their knowledge and understanding of current events to understand a post. For analysis of popular trends, the specific problem is that it is difficult to build a lexicon when the invention of new ways to refer to a word or concept is easy and common. For natural language processing in general, we argue in this paper that new uses of language in social media will challenge machines' abilities to operate with words as the basic unit of understanding, not only in Chinese but potentially in other languages.**

## I. Introduction

We propose the *pointillism* model for natural language processing. In the proposed model, a corpus is divided into grams (*e.g.*, bigrams and trigrams in the applications of this model we have developed for Chinese so far), and words and phrases are constructed from grams using external information (*e.g.*, temporal correlations in the appearance of grams). In painting, pointillism is a technique where complex scenes can be represented by arrangements of just a few primary colors into dots. This is contrasted with detailed strokes using the full pallet of colors. For machines, understanding a corpus as a time sequence of grams may be more tractable than trying to maintain an accurate lexicon for many applications and contexts. Thus, we propose the pointillism model as an alternative to natural language processing based on words.

As a specific case study, in Section II we consider the analysis of topical trends in Chinese social media. The written form of Chinese that is used for social media contains so many neologisms, creative character placements, and combinations of writing systems that it has been dubbed the "Martian Language." [1], [2]. To evade censorship, convey special meaning, or simply for creative reasons it is common for Chinese speakers to invent completely new ways to reference a topic or current event whenever they post content. This not only makes it difficult to build a lexicon that contains all possible terms, but it also exacerbates the already difficult task of Chinese text segmentation and means that the co-occurrences of words or grams alone may not be enough information to link posts about the same topic.

Figure 1 contrasts a typical traditional approach to trend analysis with the pointillism approach. In a traditional method, a lexicon is used to break the corpus up into words and then the trends of words and sequences of words are analyzed. This requires a lexicon and a good way to divide the text into words. An example of the pointillism approach might begin with trend analysis of trigrams. Then for trigrams that appear to have an interesting trend an analysis phase uses external information, such as correlations in trigram frequency over time, to build trending words and phrases out of overlapping trigrams. The main advantage of the pointillism approach is that word segmentation and other tasks that require a lexicon are not necessary, meaning that even corpuses with many unknown words can be analyzed. A disadvantage of the pointillism approach is that external information is needed, in this case the frequency of the appearance of trigrams with a relatively high resolution on the time scale. For these reasons, while a more traditional trend analysis may be more appropriate for, for example, analysis of blog posts in English [3], we assert that the pointillism approach is more effective in a different context such as the analysis of microblog posts in Chinese.

In contrast to other approaches based on grams [4], [5], the pointillism approach uses external information: detailed timing information that contains correlations based on human diurnal patterns and current events. Microblog posts have a timestamp that is accurate to the minute or second (though, in this paper, we bin posts by the hour). This timing information enables us to build words and phrases out of trigrams and then group those words and phrases into topical clusters.

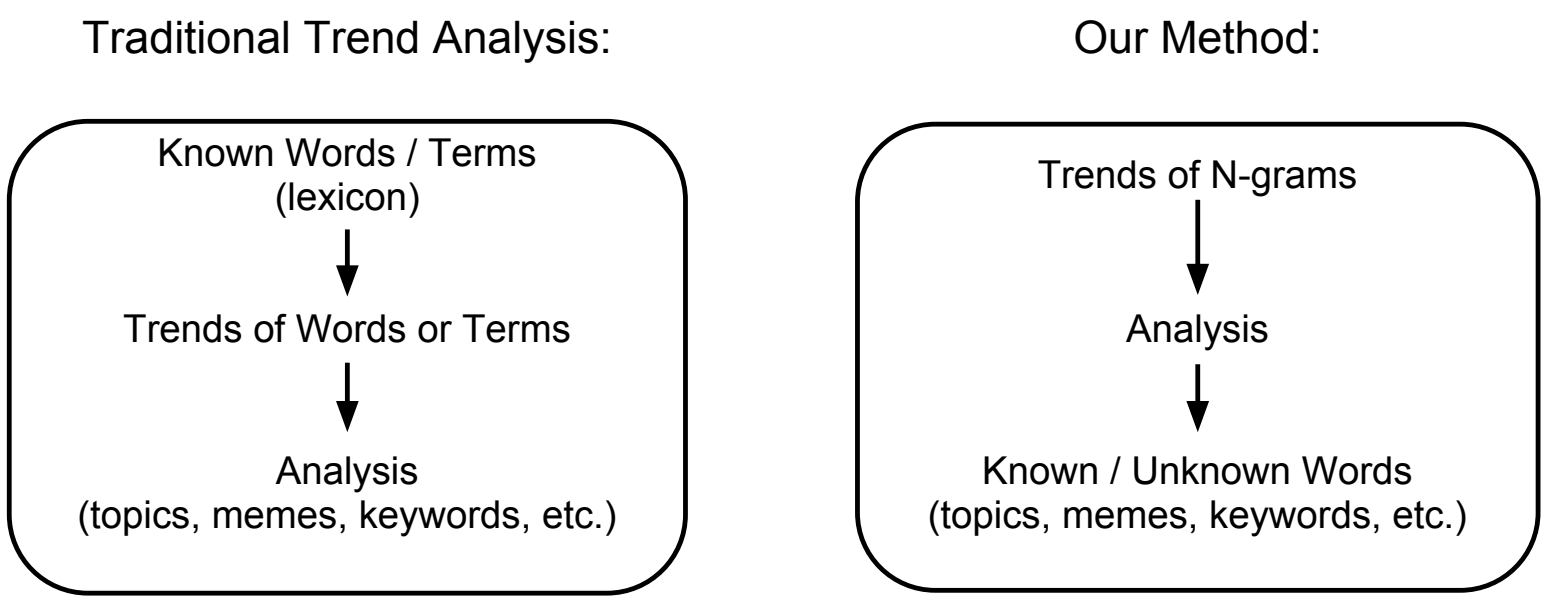

Fig. 1. Change of Order.

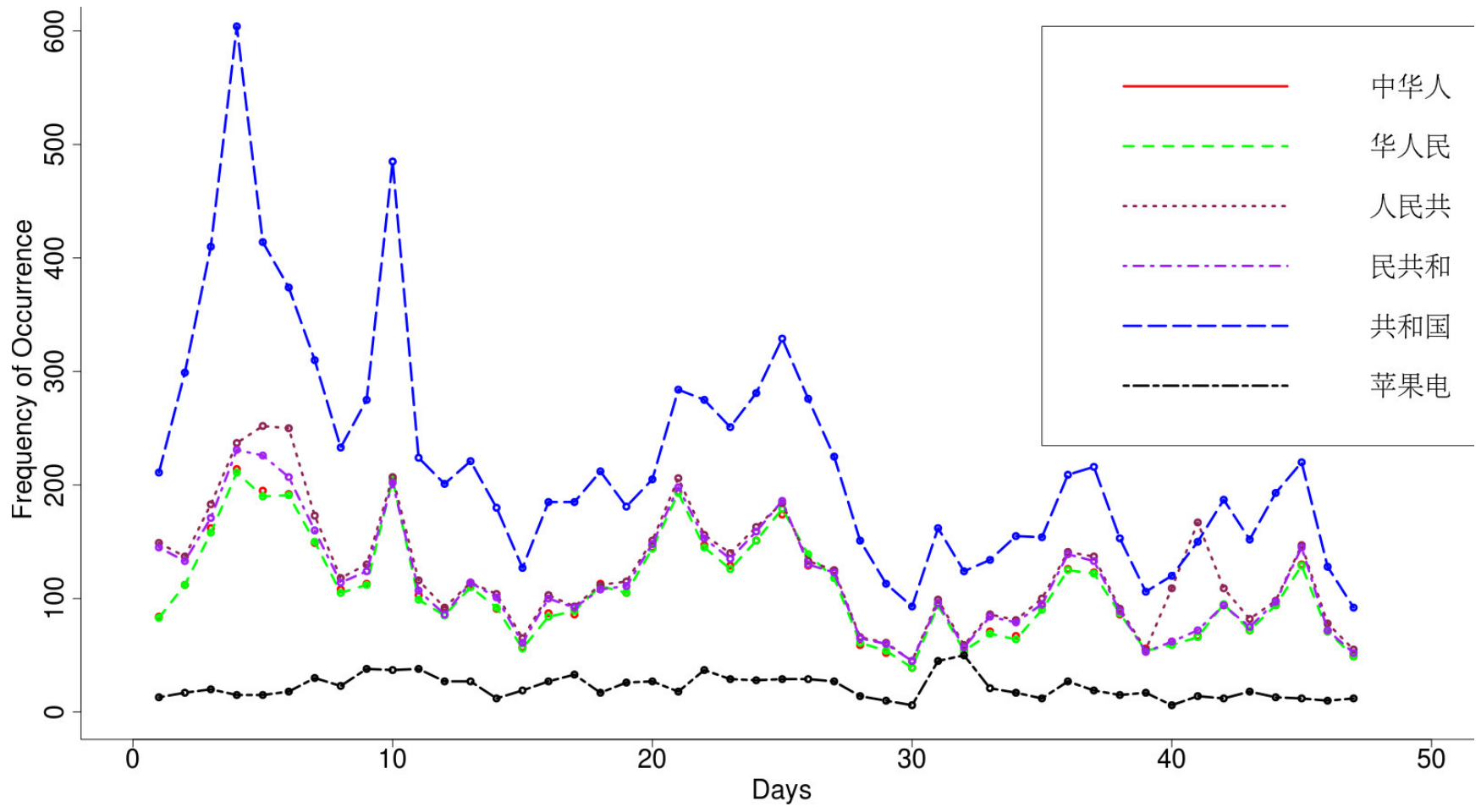

Fig. 2. **Trigram trends for 中华人民共和国 on Weibo.**

Figure 2 shows an example of the temporal patterns that we exploit. This is from a dataset that we are collecting from Weibo, a Chinese-language microblogging site that is similar to Twitter. The word we are analyzing in this figure is 中华人民共和国 (People's Republic of China). This is neither a neologism nor an unknown word that cannot be found in the dictionary, but it is a good example of a word that can create ambiguities for word segmenters. Substrings of this word are themselves words: 中华 (alternate formal name for China), 华人 (ethnic Chinese), 人民 (the people), 中华人民 (the Chinese people), and 共和国 (republic). Most individual characters in Chinese are themselves words, as well, *e.g.* 人 (person) or 和 (and). The temporal correlations shown in Figure 2 exist for unknown and uncommon words, but we use a common term in the figure so that the temporal correlations we refer to are illustrated for the reader.

The rest of this paper is structured as follows. First we present some preliminary results from an application of the pointillism approach in Section II, to demonstrate that natural language processing tasks that go beyond simple document classifications can be carried out without a lexicon. Then, in Section III, we give examples from several languages where the pointillism approach may be applicable. Then we conclude with some open questions and final remarks.

## II. Trend Analysis in Chinese Social Media

In another paper [6], we describe the detailed algorithms we used to build a system that connects trigrams of Chinese characters from Weibo together into longer words and phrases based on frequency correlations. Here we give an example from that paper where these frequency correlations can help piece together the details of a trending story even though only a meaningless trigram showed a strong trend and posts about the event shared only very common words. On 4 August 2011 the trigram 万为开 showed a strong trend. Running the algorithms described in Song *et al.* [6] to find overlapping trigrams with frequency correlations gives a set of candidates that a human can use to understand the event that lead to the trend:

```
100100\_20110804\_万为开:d: gram=万为开,
up1conn=万为开拓团掐电视,
(Wan made a TV program about the first
immigrants)
```

```
万为开拓团纪念碑被警,
(The statue was ... by the police)
万为开拓团纪念碑被泼上了,
(The statue was splashed ...)
万为开拓团纪念碑被泼红漆,
(The statue was splashed with red paint)
万为开拓团纪念碑被泼红油漆,
(The statue was splashed with red oil
paint)
万为开拓团纪念碑被5名男子,
(The statue was ... by 5 men)
万为开拓团纪念碑被5人砸,
(The statue was defaced by 5 men)
万为开拓团纪念碑被5人已离,
(The statue was ... by 5 men who have
left)
万为开拓团纪念碑被砸,
(The statue was smashed)
万为开拓团民,
(The first immigrant people)
```

万为开 is a trigram which has no meaning in Chinese. We caught this particular trigram out of the 323 million trigrams in our database because it appeared 100 times more frequently than average on 4 August 2011. After we fed this trigram into a trigram connector and set the connection time from 5 days before to 5 days afterward, we found the phrase: 万为开拓团拍电视 (Wan made a TV program about the first immigrants). This is still not clear enough to tell us why making a TV program created a trend. However, if you read the candidate results, the whole story becomes clear. It tells us that 5 men smashed and splashed red oil paint on the statue of "The First Immigrants."

In this event, there are many trigram words, such as 纪念碑 (statues), 开拓团 (immigrants), 红油漆 (red oil paint) and so on. However, the only trigram with a strong trend was, 万为开, a meaningless trigram. In general, these three characters did not appear together before this event. The sudden frequency increase of this trigram from 0 helps our system notice this trigram, which lead us to this event. Other trigrams did not increase in rate as much as 万为开 because of this event. This may be because they already exist and are common words and thus it is difficult for them to have a precipitous increase in one day.

As an example of such a connection, with an improved algorithm based on term frequency—inverse document frequency (tf-idf), we were able to make the connection between oil prices and diesel prices in two Weibo trends that were related:

```
{明天油价又要涨价了,0.9515623470222634}
(Tomorrow oil prices will again rise)
{明天油价又要涨了,0.9359352855815268}
(Tomorrow oil prices will again swell)
{明天油价上涨,0.8963352077911995}
(Tomorrow oil prices rise)
{明天油价上涨(的),0.5048456523116667}
(The rise of oil prices tomorrow)
{明天油价要,0.8970978381471049}
(Tomorrow oil prices will)
{柴油370元和0号柴油价格,0.9035492192818616}
(Diesel fuel 370.00 yuan diesel fuel price)
{柴油370元和0号柴油价格(上),
0.7128173787969483}
(Diesel fuel 370.00 yuan diesel fuel price
above}
{柴油370元和0号柴油0,0.8553821969195776}
(Diesel fuel 370.00 yuan price)
{柴油370元和0号汽油390,
0.8398579004604106}
(Diesel fuel 370.00 yuan gas 390)
{柴油390,0.8944413356551058}
(Diesel fuel 390)
```

The algorithms and complete results will be published in a future paper.

## III. Social media examples

In this section, we give some examples in various languages where approaches based on lexicons may fail.

### A. An example from Japan

The following example of a post from Twitter is a mixture of English, Kanji (Chinese characters), and Hiragana (a Japanese syllabary):

三日更衣室で今北postしますね

A rough translation of the post is, "I have been in the dressing room for three days, can you make a post to explain in three lines what has been going on?" "三日更衣室で" is relatively straightforward Japanese text that translates to, "I have been in the dressing room for three days." "今北" is a neologism that is explained in a blog post by Maggie Sensei [7]. "今北" is the name of a company (Imakita), but in this case serves as a neologism for "今、来た 三行", which translates to "I just got here, explain in three lines what has been going on." " " in Japanese is pronounced "Sangyou," but sounds like "三行", or "sanggyou", meaning "three lines." "しますね" is Japanese Hiragana for "make", and "post" is an English word.

This example demonstrates the ability of Japanese social media users to mix different character systems, play with sounds, and combine different languages. This gives microblog posters great freedom of expression in how they use written language to express their meaning. When the posters are using this freedom while posting about current events and hot trends, natural language processing based on a lexicon may not be able to categorize the topic of the post based only on the words in the lexicon.

### B. An example from India

It is very common in India to mix languages in social media posts, such as this example from Twitter that uses a mixture of Hindi and English:

**#saale government ki #maakiaankh #kutteke bacche social networking sites bandh karega ek ek #madharchod ka #gaand maro #bhenchod**

This translates to:

**#Bad-boy government's #eye-of-mother #son-of-dogs want social networking sites shut down each and every #motherfucker's #ass be raped #sisterfucker**

A topical analysis of social media aimed at sentiment analysis would be concerned with the topic of the post that the sentiment is being directed at. In this case it is an alleged government proposal to shut down social media sites. In this post, "government" and "social media sites" are in English but the words expressing the desire to ban these sites and the author's sentiment towards this are in Hindi. Indian social media also contains many regional languages, so any sentiment analysis of Indian social media sites would have to incorporate many different ways of mixing languages. Can the pointillism approach help in this case, or is it better to build build a comprehensive lexicon that incorporates these many languages? We leave this as an open question.

### C. *An example from China*

The following is an excerpt from a blog post in China:

雅青别说我粗暴，这样说也是没法子的事，现在的形势就是复读机，都可以填上平仄: 中南海管得了南海，钓鱼台管得了钓鱼岛

The author is using poetry and creatively pointing out character similarities in different words to make a point without explicitly writing their meaning. The phrase is a poem and translates to, "Yaqing don't call me rude, this kind of talk is also at a loss for meaning, the circumstances now are a broken record, that can be expressed with a poem: Zhongnanhai has authority over the South China Sea, Diaoyutai State Guesthouse has authority over the Diaoyu islands." Zhongnanhai is the central headquarters of China's State Council, and Diaoyutai State Guesthouse is a hotel where leaders of foreign nations stay when they visit Beijing. The author is expressing an opinion about a recent event in which China demanded an apology from Japan over the Diaoyu island dispute rather than demanding compensation. "雅青" is a neologism that refers to "Elegant Youth" in contrast to "愤青", which means "angry youth." In this context "雅青" means people who would prefer a less militant solution to the Diaoyu Island dispute. "复读机" is a repeating machine that records audio and then plays it back (*e.g.*, for use as a study aid in the classroom to record lectures for later review), so in this context is similar to the English phrase "broken record" in terms of connotations.

When authors use this kind of indirect language to express opinions about current events, the co-occurrences of known words and individual grams sometimes are not enough to adequately capture the full range of topics touched on by a post. The trigram "钓鱼岛" suggests that the example post is related to Diaoyu island which was probably a trend at the time the post was written, but the post also has phrases that suggest government authority ("中南海管得了"), foreign influence ("钓鱼台管得了"), a situation ("现在的形势"), and circumstances that have not changed ("复读机"). Other authors will have other methods for expressing these concepts. An advantage of the pointillism approach is that external information, such as frequency correlations in trigrams, can help to link together similar concepts that are expressed in very different ways.

Humans are able to use their understanding of current events to understand the meaning of a blog post such as the example above. For natural language processing, a machine lacks this contextual information but is able to perform calculations on how the trigrams that might be parts of words or phrases correlate over time in terms of the frequency of their usage. This information can serve as proxy information for context and may enable natural language processing tasks to make connections that go beyond co-occurrences and get closer to the actual meaning of social media content.

## IV. Conclusion

In conclusion, we have given supporting evidence to suggest that carrying out a natural language processing task, specifically trend analysis, is possible without a lexicon is possible. We have also given examples in several languages where removing the dependence on a lexicon could be beneficial. Thus we propose the pointillism approach to natural language processing as being particularly suitable for certain languages and contexts.

## Acknowledgment

We sould like to thank the anonymous reviewers for valuable feedback. This material is based upon work supported by the National Science Foundation under Grant Nos. #0844880, #0905177, #1017602. Jed Crandall is also supported by the Defense Advanced Research Projects Agency CRASH program under grant #P-1070-113237.